# Unsupervised Image Translation using Adversarial Networks for Improved Plant Disease Recognition

Haseeb Nazki
Department of Electronics Engineering
Chonbuk National University
Jeonju, South Korea

Sook Yoon
Department of Computer Engineering
Mokpo National University
Muan, South Korea

Alvaro Fuentes
College of Artificial Intelligence
Tianjin University of Science and Technology
Tianjin, China

Dong Sun Park
IT Convergence Research Centre
Department of Electronics Engineering
Chonbuk National University
Jeonju, South Korea

**Abstract**

Acquisition of data in task-specific applications of machine learning like plant disease recognition is a costly endeavor owing to the requirements of professional human diligence and time constraints. In this paper, we present a simple pipeline that uses GANs in an unsupervised image translation environment to improve learning with respect to the data distribution in a plant disease dataset, reducing the partiality introduced by acute class imbalance and hence shifting the classification decision boundary towards better performance. The empirical analysis of our method is demonstrated on a limited dataset of 2789 tomato plant disease images, highly corrupted with an imbalance in the 9 disease categories. First, we extend the state of the art for the GAN-based image-to-image translation method by enhancing the perceptual quality of the generated images and preserving the semantics. We introduce AR-GAN, where in addition to the adversarial loss, our synthetic image generator optimizes on Activation Reconstruction loss (ARL) function that optimizes feature activations against the natural image. We present visually more compelling synthetic images in comparison to most prominent existing models and evaluate the performance of our GAN framework in terms of various datasets and metrics. Second, we evaluate the performance of a baseline convolutional neural network classifier for improved recognition using the resulting synthetic samples to augment our training set and compare it with the classical data augmentation scheme. We observe a significant improvement in classification accuracy (+5.2%) using generated synthetic samples as compared to (+0.8%) increase using classic augmentation in an equal class distribution environment.

**Keywords:** Generative Adversarial Networks; Image Translation; Data Augmentation; Convolutional Neural Network; Plant Disease Recognition; Class Imbalance; Activation Reconstruction loss

## 1. Introduction

Plant pests and diseases affect food crops and hence play a considerable role in the annual yield and economic losses in the agricultural sector, threatening the food security of a country (Strange et al., 2005). For instance, during the ages, tomato production has been experiencing a considerable increase worldwide, hence representing the most economically important vegetable worldwide (Bergougnox, 2014). However, tomato plants are vulnerable to numerous syndromes and attacks caused by diseases and pests. There are several reasons that contribute to these effects on the crops: (1) environmental conditions, such as humidity, temperature, light, nutritional excess (fertilizer), and species of plant constituting the abiotic disorders; (2) pests, such as whiteflies, leaf miners, worms, bugs, etc. that inhabit and colonize from plant to plant; and (3) bacteria, viruses, and fungi that constitute to the most common diseases (Bergougnox, 2014). These diseases and pests show different physical characteristics with the plant, such as changes in shape, color, form, etc.

of distinct parts in the plant. Therefore, these variations are hard to be distinguished, which furthermore makes their recognition a challenge for an early detection and treatment that could help avoid several losses in the whole crop. These effects are becoming a demanding menace nowadays and need to be approached in the best way and with special attention. Furthermore, monitoring the plants for the symptoms continuously, can prove to be a tedious task. Hence, efforts have been devoted to come up with the approach that could automate the process of recognition of diseases using image data. An early recognition of diseases and a proper treatment in due course can be helpful in dealing with huge damages and improve the crop yield.

In recent years, machine learning-based methods have attained state-of-the-art performance in many computer vision tasks. Hence, several of the recent approaches towards this plant disease recognition challenge are based on training a deep neural network associated with a classifier for the identification of diseases using the images of the plant parts (Akhtar et al., 2013; Al-Hiary et al., 2011; Brahimi et al., 2017; Atabay et al., 2017). One of the advantages of using deep neural networks is its capability to exploit the raw data directly without using the hand-crafted features. The success of machine learning is largely related to the access to enormous amounts of data for training an algorithm and the high computing power provided by graphics processing units (GPUs) which makes it possible to train these deep neural networks and enforce the parallelism in data computing. Compared to shallow learning methods which include support vector machines (SVM), decision trees and naïve bayes, deep learning models exhibit more representative power by passing input data through several non-linearity functions to produce robust descriptive features and perform recognition based on those features.

One of the main challenges in machine learning related to the agricultural imaging domain is how to cope with the small datasets and limited number of annotated samples, specifically when employing supervised machine learning algorithms that need labeled data and large number of training examples. Collecting plant disease related data is a complex and expensive procedure and requires the collaboration of people from different fields at contrasting stages. Although public datasets are available, most datasets are still limited in size and applicable to specific problems. Also, researchers often come across the challenge of class imbalance which has been a general problem in machine learning. Using classic data augmentation for enlarging training set and balancing classes has been reported in various literatures (Perez et al., 2017; Wong et al., 2016). However, the diversity and variation that can be gained from such modifications of the images (such as rotation, translation, flip and scale) is relatively small. This motivates the use of synthetic data, where the generated samples introduce more variability and can enrich the dataset further, to improve the recognition training process and accuracy.

Generative Adversarial Networks (GANs) introduced by Goodfellow et al., (2014), have been found successful at various tasks of generating synthetic images. The major goal is to generate synthetic samples with the same characteristics as the given training distribution. Motivated by the success, GANs used in image-to-image translation i.e. translation of one possible representation of a scene to another, further proves its representational power (Isola et al., 2017; Zhu et al.,2017; Yi et al., 2017). Though, existing state-of-the-art methods (Zhu et al., 2017; Yi et al., 2017) eliminate the need for paired data for image-to-image translation, the generated images often depict visible flaws that lack structural definition for an object of interest.

We begin by extending the state of the art for GAN-based image-to-image synthesis to improve the perceptual quality of the generated images by preserving the structure of the scene. The main reason to use GANs in an image translation setting is the effective synthesis of viable output, given a limited amount of input data. Our approach builds on CycleGAN and DualGAN (Zhu et al., 2017), where they propose a self-consistency or reconstruction loss that preserves the input image after the translation cycle. Though, these approaches combine the adversarial loss and the cycle-consistency loss (Zhu et al., 2017) to preserve the semantic affinity and avoid mode collapse but an explicit regularization term that could penalize the perceptual quality between the images from the two domains is absent.

In this work we introduce AR-GAN that differs from the previous approaches by optimizing on an activation reconstruction loss (Johnson et al., 2016; Cha et al. 2017) in addition to regularizing the original GAN objective function and cycle-consistency optimizations to present visually more compelling synthetic images on an unaligned dataset. The main focus of this work is to analyze the performance of plant disease recognition systems using synthetically generated image data. We synthesize high-quality plant disease samples from a finite number of raw images using AR-GAN. We further use these synthetic samples for augmentation and balancing the data in the training set. As shown in Fig. 1, our pipeline feeds the generated samples into another learning machine (i.e. a Convolutional Neural Network). Finally, we compare the classical and synthetic data augmentation schemes using various metrics. To the best of our knowledge, this is the first work that uses GANs to synthetically augment the dataset to improve the plant disease recognition performance. Our main contributions are as follows:

- We formulate unsupervised AR-GAN to learn one-to-one mapping for high quality image translation by aggregating cycle-consistency and activation reconstruction loss.

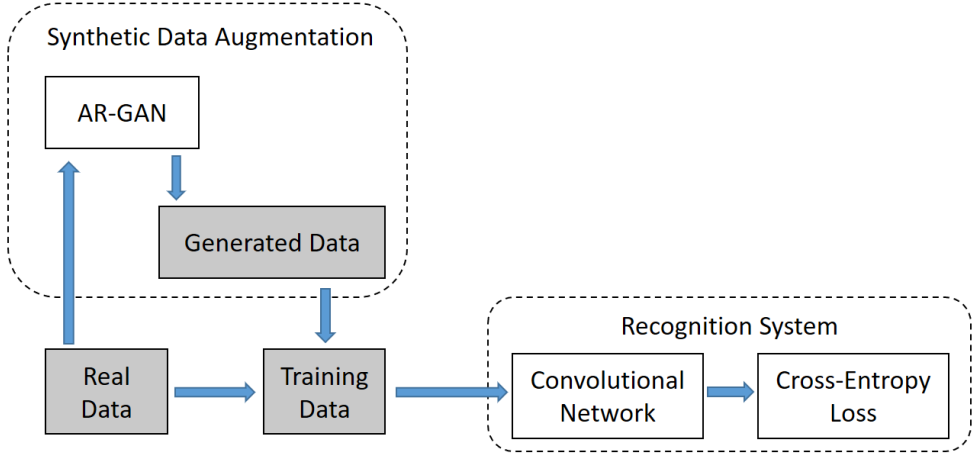

**Fig.1.** Our proposed pipeline containing two components: a Synthetic Data Augmentation module with AR-GAN for unsupervised learning and a Recognition System with a Convolutional Neural Network for supervised learning. We aim to learn more discriminative embeddings with the "training data" containing both "real data" and "generated data".

- We propose a simple pipeline for synthetic augmentation of plant disease datasets using AR-GAN to improve the plant disease recognition performance in a data deficient environment.
- We introduce our limited dataset of tomato plant disease images to validate the effectiveness of our pipeline to prove or disprove the hypothesis: (i) Does synthetic augmentation improve the performance of a deep convolutional neural network for plant disease recognition? (ii) How does synthetic augmentation compare to classic augmentation in terms of performance in a plant disease recognition system?

## 2. Materials and Methods

### 2.1. Data Augmentation and Generation

The biggest limitation with machine learning algorithms is that they require huge amounts of training data before they become effective. One of the more common and effective ways to obtain more data is classical data augmentation (Perez et al., 2017). There are multiple ways to augment images, but commonly used ones include traditional transformations like flipping images horizontally, flipping images vertically, random crops, zooms, rotations, color perturbation, and translation. Interestingly, augmentation in data-space using elastic deformations has been reported to give better improvement in error % than augmentation in feature space (Wong et al., 2016) in some fields like medical images.

Generative network approaches have been extensively used to generate samples in recent years. (Kingma et al., 2013; Sohn et al., 2015; Mirza et al., 2014; Arjovsky et al., 2017; Goodfellow et al., 2014; Yi et al., 2017) have produced nice samples on various image datasets (Radford et al., 2015; Nguyen et al., 2017; Yi et al., 2017). All the generative models $G$ are parameterized by $\theta$ and take as input a random noise $z$ and output a sample $G(z; \theta)$, such that the output can be considered as a sample drawn from a distribution: $G(z; \theta) \sim p_g$. The training objective for the generative model $G$ is to approximate $p_d$ using $p_g$, given the training data $x$ drawn from $p_d$. We limit our discussion to GANs in the next section. The main reason for using GAN based image translation in our study is that in practice, other generative models tend to produce blurry images relative to GANs.

### 2.2. Generative Adversarial Networks (GANs)

GANs (Goodfellow et al., 2014) consist of two separate neural networks: a generator $G$ that takes a random noise vector $z$ and generates synthetic data $G(z)$; a discriminator $D$ that takes an input $x$ or $G(z)$ and output a probability $D(x)$ or $D(G(z))$ to indicate whether the input derived from the synthetic distribution of $G(z)$ or from the true data distribution.

GANs (Mirza et al., 2014; Arjovsky et al., 2017; Goodfellow et al., 2014) are a framework of models that learn by a two-player game between two different networks: a generator that learns to produce images from a distribution $p_d$ and a discriminator that learns to discriminate between the generated and the real images. The generator wants to fool the discriminator and the discriminator wants to beat the generator. The value function devised to be optimized is represented as follows:

$$\min_G \max_D V(D, G) = \mathbb{E}_{x \sim p_d(x)}[\log D(x)] + \mathbb{E}_{z \sim p_z(z)}[\log(1 - D(G(z)))] \quad (1)$$

where $p_d(x)$ denotes the true data distribution and $p_z(z)$ denote the noise distribution. If the discriminator is trained much better than the generator, it can discard the samples from generator with a high confidence (close to 1), and thus the loss $log(1 - D(G(z)))$ would saturate and $G$ would not learn anything from zero gradient. To avoid this, instead of training $G$ to minimize $log(1 - D(G(z)))$, we train it to maximize $log D(G(z))$. This new adversarial loss function for $G$ provides the same direction of gradient and does not saturate (Goodfellow et al., 2014).

The first model (Goodfellow et al., 2014) proposed, uses fully connected layer as its building block. More recent approaches (Radford et al., 2015; Dumoulin et al., 2016) successfully demonstrated the use of fully convolutional neural networks to achieve better performance, and since then convolution and transposed convolution layers (Dumoulin et al., 2016) have become the fundamental components in many models.

While GAN is known to be very effective in image synthesis, its training process is very unstable and requires following many guidelines to obtain satisfying results (Radford et al., 2015; Goodfellow et al., 2014). GAN also suffers from the problem of mode collapse (Radford et al., 2015; Denton et al., 2015; Goodfellow et al., 2014)] where all points collapse to a common single space. Many methods have been proposed to address this problem (Salimans et al., 2016; Che et al., 2016; Donahue et al. 2016). WGAN (Arjovsky et al., 2017) proposes to use the Wasserstein distance to measure the similarity between true and the learned data distribution, instead of using Jensen-Shannon divergence as in the original GAN model (Goodfellow et al., 2014). While it theoretically avoids mode collapse, it results in the model to take longer time to converge than previous approaches. To avoid this problem, WGAN-GP (Gulrajani et al., 2017) suggests using gradient penalty, instead of weight clipping used in WGAN. It produces better images and significantly avoids mode collapse and is also easy to apply to the training framework of other models. GANs have demonstrated potential in generating images for specific fields like image in-painting, image super resolution, text-to-image synthesis, image-to-image translation (Creswell et al., 2018). In our work, we do not focus on investigating all these sophisticated GAN algorithms. Instead, we use GANs for image translation to generate samples from the training data.

### 2.3. Image-to-Image Translation

Image-to-Image translation is defined as the task of translation of one possible representation of a scene to another, such as mapping grayscale images to RGB or the other way around (Isola et al., 2017). GANs have been used for image-to-image translation in both *supervised* as well as *unsupervised* settings. By supervised we mean that there is a corresponding ground-truth image in the target domain (Isola et al., 2017). Our approach builds on unsupervised image-to-image translation GAN architectures where they propose a self-consistency or reconstruction loss that preserves the input image after the translation cycle (Zhu et al., 2017; Yi et al., 2017; Kim et al., 2017). They all share the same framework with two generators ($G_{AB}$ and $G_{BA}$) doing opposite transformations, which can be seen as a kind of dual learning (He et al., 2016) and two discriminators $D_A$ and $D_B$ that predict whether an image belongs to that domain or not. Among recent models that use unpaired data (Aytar et al., 2017; Liu et al., 2016) a weight-sharing strategy to learn representation across domains is reported. Concurrently, other works (Taigman et al., 2016; Shrivastava et al. 2017; Bousmalis et al., 2017) encourage the input and output to share some specific feature contents. These methods use GANs with additional terms to enforce a close association between input and output in terms of image pixel space, class label space, and image feature space.

### 2.4. Plant disease recognition

Plant diseases recognition is a critical topic that is being studied through the years and is motivated by the requirement of a healthy agricultural output (Nutter et al., 2016; Bergougnox, 2014; Strange et al., 2005). The research associated can be broadly categorized as (i) Shallow learning approaches that use hand crafted features (ii) Deep learning approaches using deep convolutional neural networks to learn the feature representations. In disease recognition, shallow learning approaches have been applied for classifying tomato powdery mildew against healthy leaves by means of thermal and stereo images (Prince et al., 2015), detecting yellow leaf curl virus in tomatoes by using a set of classic feature extraction steps, classified by SVM (Mokhtar et al., 2015), recognition of tomato diseases in a greenhouse (Chai and Wang, 2013) etc. Moreover, there has been a lot of research going on in this field using plant leaf analysis together with machine learning (Singh et al., 2016). Deep learning approaches were introduced in plant image recognition as its first application came up with the identification of plant based on leaf vein patterns (Grinblat et al., 2016). They classified three leguminous plant species: white bean, red bean and soybean using CNNs and experimented using 3-6 layers. Ghazi et al., (2017) used three popular CNN architectures, AlexNet (Krizhevsky et al., 2012), VggNet (Simonyan and Zisserman, 2014), and GoogLeNet (Szegedy et al., 2015), and evaluated the numerous factors affecting the performance of these networks on task of plant species identification. They use Transfer learning to fine-tune the pre-trained models, using LifeCLEF plant dataset (Goëau et al., 2014) and applied classic data augmentation techniques based on image transforms such as rotation, translation, reflection, and scaling to decrease the chance of overfitting.

Other applications of the deep CNNs in plant disease recognition include the diagnosis of crop leaf disease in (Mohanty et al., 2016), where they used a deep convolutional neural network to identify 26 diseases and 14 crop species using a publicly available PlantVillage dataset (Hughes et al., 2015) with over 50,000 images. Among the recently published works in the literature [(Brahimi et al., 2017; Atabay et al., 2017)], they use transfer learning of pre-trained models and introduce deep residual learning (He et al., 2016) to identify tomato plant diseases of the PlantVillage dataset.

## 2.5. Network Overview and Architecture

As mentioned earlier, the goal of our proposed method is to observe the improvement in the accuracy of plant disease recognition systems suffering with class imbalance and sample deficiency in the training data. To enlarge the training dataset while keeping classes balanced, the system requires an additional data augmentation measure. We propose to synthetically generate additional training data using GAN and further train the recognition network using that data in addition to the original data. Further, we also investigate the effectiveness of the synthetic data augmentation by GAN over classical data augmentation for improved recognition.

In this section we elaborate on the pipeline of the proposed method. As shown in Fig. 1, our proposed system consists of two blocks: a) Synthetic Data Augmentation block using AR-GAN for data synthesis and class balancing b) Recognition System that uses a deep convolutional neural network for image recognition. The real data is used to train the AR-GAN model which will generate more training data. The generated data from AR-GAN is added to the real data to yield training data for the CNN. In the following section, we introduce the architecture for our AR-GAN, followed by the baseline CNN architecture used to analyze the recognition performance.

### 2.5.1. AR-GAN

A framework of our AR-GAN is depicted in Fig. 2. The AR-GAN improves on CycleGAN (Zhu et al.,2017), unsupervised image-to-image translator, by introducing an activation reconstruction module consisting of a feature extraction network to calculate Activation Reconstruction Loss (ARL) between an image and its translation, in addition to the cycle-consistency loss and adversarial loss of CycleGAN.

The objective of unsupervised image-to-image translation is to unveil the mappings between two distinct domains $A$ to $B$ and vice versa, using unpaired samples from distributions $p_d(a)$ and $p_d(b)$ in each domain. This can be expressed as a conditional generative modeling task where we try to approximate the true conditionals $p(a|b)$ and $p(b|a)$ using samples from the true marginals. A significant postulation here is that elements in domains $A$ and $B$ are highly correlated; otherwise, it is questionable that the model would uncover a meaningful relationship without any pairing information. The CycleGAN model (Zhu et al.,2017) approximates these conditionals using two mappings $G_{AB}: A \rightarrow B$ and $G_{BA}: B \rightarrow A$, parameterized by deep networks, which satisfy the following constraints: 1) The output of each mapping must match the empirical distribution of the target domain, when focused over the source domain. 2) Mapping an element from one domain to the other, and then back, should deliver a sample close to the element that we initially started with, thus showing a cyclic-consistency. The first constraint is satisfied using the GAN (Goodfellow et al., 2014). Applying this matching on target domain $B$, marginalized over source domain A, includes minimizing an adversarial objective with respect to $G_{AB}$:

$$\mathcal{L}_{GAN}(G_{AB}, D_B) = \mathbb{E}_{b \sim p_{B(b)}}[\log D_B(b)] \quad (2)$$
$$+ \mathbb{E}_{a \sim p_{A(a)}}[\log(1 - D_B(G_{AB}(a)))]$$

as the discriminator $D_B$ is trained to maximize it. A parallel adversarial loss $\mathcal{L}_{GAN}(G_{BA}, D_A)$ is defined in the opposite direction.

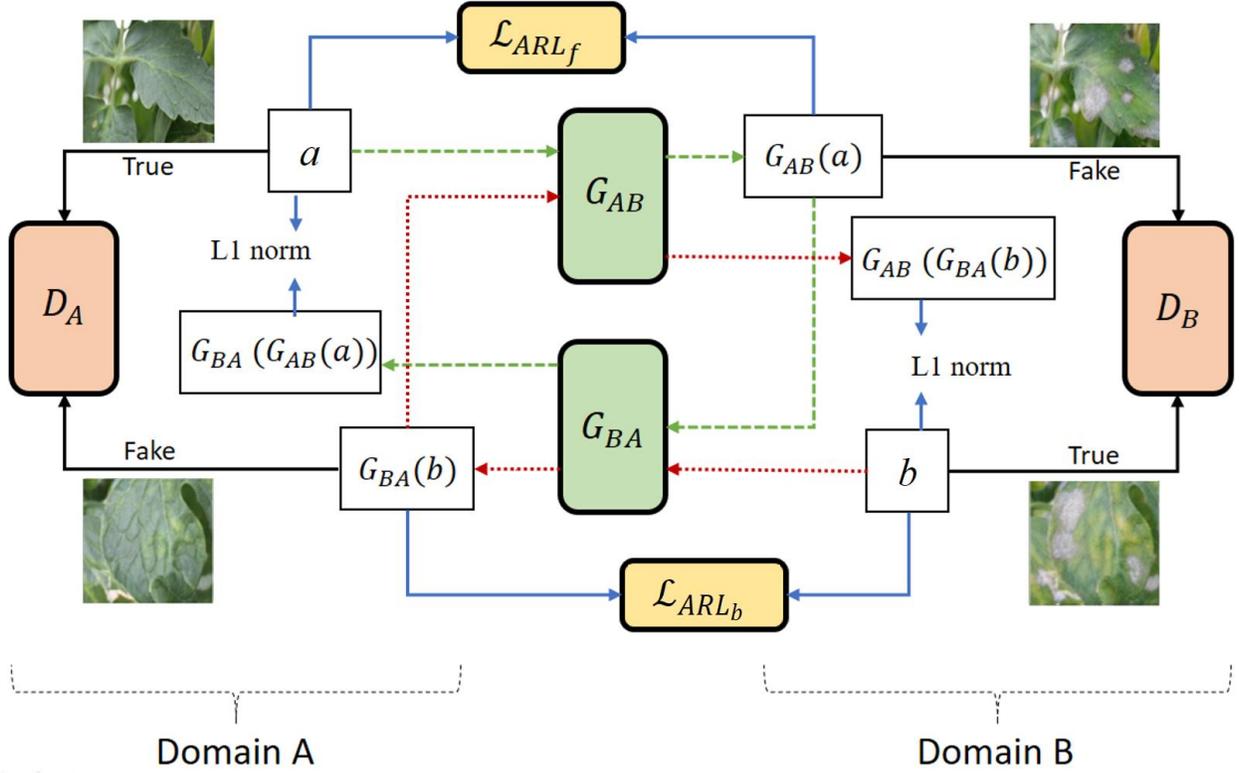

**Fig. 2.** Framework of AR-GAN. A and B are two unaligned domains. Two generators ($G_{AB}$, $G_{BA}$) translate an image from one domain to another. There is a discriminator for each domain ($D_A$, $D_B$) that judges if an image belongs to that domain. Two cycles of data flow, the green one performs a domain transfer A → B → A, while the red one is B → A → B. L1 loss is applied on the input a (or b) and the reconstructed input $G_{BA}(G_{AB}(a))$ (or $G_{AB}(G_{BA}(b))$) to enforce self-consistency. In addition to that, a common feature extraction network calculates activation reconstruction loss between $a$ and $G_{AB}(a)$: $\mathcal{L}_{ARL_f}$ or $b$ and $G_{BA}(b)$: $\mathcal{L}_{ARL_b}$ or both.

Cycle-consistency imposes that, when initializing with a sample "$a$" from $A$, the reconstructed $a' = G_{BA}(G_{AB}(a))$ remains close to the original "$a$". For image domains, closeness between $a$ and $a'$ is typically measured with $L1$ or $L2$ norms. When using the $L1$ norm, cycle-consistency starting can be formulated as:

$$\mathcal{L}_{cyc}(G_{AB}, G_{BA}) = \mathbb{E}_{a \sim p_{A(a)}} \| G_{BA}(G_{AB}(a)) - a \| \quad (3)$$
$$+ \mathbb{E}_{b \sim p_{B(b)}} \| G_{AB}(G_{BA}(b)) - b \|$$

Finally, the full objective is given by:

$$\mathcal{L}_{cycleGAN}(G_{AB}, G_{BA}, D_A, D_B) \quad (4)$$
$$= \mathcal{L}_{GAN}(G_{AB}, D_B)$$
$$+ \mathcal{L}_{GAN}(G_{BA}, D_A)$$
$$+ \alpha \mathcal{L}_{cyc}(G_{AB}, G_{BA})$$

where α is a hyper-parameter that stabilizes the two constraints. The adversarial terms encourage generation of realistic samples in either domain while cycle-consistency ensures a strong relationship between these domains.

We adopt the activation reconstruction loss $\mathcal{L}_{act}$ (Johnson et al., 2016; Cha et al. 2017) as a perception loss trained along with the other loss functions aiming to enforce perceptual realism between the real and the generated images and increasing the stability of the model. $\mathcal{L}_{act}$ encourages high-level feature representations of the images to be alike. Let $A_a^n$ and $A_b^n$ be the activation outputs of the nth layer within any convolutional recognition network (feature extraction network) where $a$ and $b$ are used as inputs respectively. Then $\mathcal{L}_{ARL}$ is defined as:

$$\mathcal{L}_{ARL} = \frac{1}{m} \|A_a^n - A_b^n\|_F^2 \qquad (5)$$

where $\|\cdot\|_F$ represents the Frobenius norm, $m$ is the shape of the feature map and $n$ is $n$th layer used from the feature extraction network. From equations (4) and (5) our total objective function can be summed up as:

$$\mathcal{L}_{total} = \mathcal{L}_{cycleGAN}(G_{AB}, G_{BA}, D_A, D_B) + \lambda \mathcal{L}_{ARL} \qquad (6)$$

where $\lambda$ is the hyper-parameter to regulate the $\mathcal{L}_{ARL}$ term. The resemblance to the new domain and faithfulness to the original image is a trade-off determined by the weight $\lambda$ of the ARL term relative to the image adversarial term. If $\lambda$ is set too large, the translated images are close to the input but cannot capture the features of the other domain and if $\lambda$ is set too small, the translated images fail to pertain the perception value of the input. In our experiments, we start by setting $\lambda = 0$, which means we only use the cycle-adversarial constraint to train the generator. Then we gradually increment $\lambda$ which leads to the increase in the aesthetic and perceptive quality until a threshold is reached. The value of $\lambda$ is largely a domain dependent hyper-parameter.

### 2.5.1.1. Architecture

The architectures for the generator and discriminator are similar to (Johnson et al., 2016; Zhu et al., 2017). $G_{AB}$ and $G_{BA}$ contain 9-residual blocks, pair of convolution block with stride 2 and another pair of fractionally strided convolution block with stride 1/2. For $D_A$, $D_B$ we use 70 × 70 PatchGAN (Isola et al., 2017; Zhu et al., 2017) discriminator, which aims at classifying whether 70 × 70 overlapping image patches are real or fake. Such a discriminator architecture working at a patch-level, has fewer parameters than any full-image discriminator and can work on arbitrarily-sized images in a fully convolutional fashion (Isola et al., 2017). The feature extraction network, to calculate $\mathcal{L}_{ARL}$, consists of a down convolution layer followed by 4-residual blocks.

Let: -
- C7-k-n: 7×7 Convolution Instance Norm ReLU layer with n filters and stride k.
- D3-2-n: 3×3 Convolution Instance Norm ReLU layer with n filters and stride 2.
- R3-n: Residual block that contains two 3 × 3 Convolutional layers with the same n number of filters on both layers.
- U3-n: 3×3 fractional strided Convolution Instance Norm ReLU layer with n filters and stride 1/2.
- P4-2-n: 4 × 4 Convolution Instance Norm Leaky ReLU layer with n filters and stride 2.

Following (Zhu et al., 2017) the Generator network with 9 blocks consists of:
*C7-1-32, D3-2-64, D3-2-128, R3-128, R3-128, R3-128, R3-128, R3-128, R3-128, R3-128, R3-128, R3-128, U3-64, U3-32, C7-1-3*

The discriminator architecture is:
*P4-2-64, P4-2-128, P4-2-256, P4-2-512*
where we again follow [(Zhu et al.,2017)] and for *P4-2-64* circumvent instance normalization.

The feature extraction network to calculate $\mathcal{L}_{ARL}$ is given by:
*C7-2-64, R3-64, R3-128, R3-256, R3-512*

The feature extraction network is warm-started with pre-trained ImageNet weights. We did some extensive experiments to try different combinations of feature layers to obtain the best results. Using all layers ensures the preservation of low-level as well as high level traits resulting in best performance. The network architecture trains on and outputs images of size 256×256. For calculating $\mathcal{L}_{cycleGAN}(G_{AB}, G_{BA}, D_A, D_B)$, we use least squared loss instead of log likelihood for generating higher quality results. Weights were initialized from a Gaussian distribution with 0 mean and 0.02 standard deviation. We train for 200–300 epochs for different datasets where we keep the same learning rate for the initial 100 epochs and linearly decay this learning rate to zero over the next remaining epochs. For most of the experiments with Tomato Plant Disease Dataset, we find it suitable to use α =10 in equation (4) and $\lambda = 1$ in equation (6), which are the terms that largely depend on the nature of the two domains. All models are trained within 3 days on a single NVIDIA Titan X GPU with Adam optimizer and a batch size between 1 and 3. The network has approximately 11.3M parameters and the generation of synthetic images takes about 0.0182 ms per image. We implemented our generative model using the Pytorch framework.

### 2.5.2. CNN for Plant disease Recognition

CNNs are extensively used for solving image recognition tasks in computer vision. CNN architectures for tomato plant disease recognition used in (Sladojevic et al., 2016; Lu et al., 2017; Brahimi et al., 2017) usually either contain few convolution layers because of the small datasets or many convolution layers with large datasets. We use a simple baseline Resnet50 model for various experiments to analyze the effectiveness of classical and synthetic augmentation techniques using different instances of our tomato plant disease dataset. The model takes in fixed 3-channel input RoI of $256 \times 256$. The architecture remains the same, consisting of 34-layers of 3-layer bottleneck block where each block contains $1 \times 1$, $3 \times 3$, and $1 \times 1$ convolutions. The layers with $1 \times 1$ convolution are responsible for reducing and then restoring (increasing) the dimensions, leaving the 3×3 layer as a bottleneck with smaller input/output dimensions (He et al., 2016). We use ReLU as activation functions and with the network having approximately 23M trainable parameters. We fine-tune our recognition networks from ImageNet pre-trained weights (Chai and Wang, 2013). For training, we use a batch size of 32 with a learning rate of 0.001 for 300 epochs. We use stochastic gradient descent optimization with Nesterov momentum updates (Nesterov, 1983). For the implementation of the recognition CNN architecture we used the Keras framework.

## 3. Results

We first introduce our limited tomato plant disease recognition dataset along with other datasets used to evaluate the performance of our GAN network. We compare our AR-GAN against recent methods for unpaired image-to-image translation on both paired as well as unpaired datasets. We study the importance of ARL term and compare our final method against several variants. Finally, we evaluate the proposed pipeline on our tomato plant disease dataset. The main challenge here is to manage with the small amount of available data for training the final classifier network. This setup can be imitated by other applications facing the problem of discrete training data e.g. medical images.

### 3.1. Datasets

In this work, we evaluate in terms of two tasks a) our generative model and b) the recognition system of the proposed method. We experiment on several datasets. To evaluate the performance of our generative model we use Cityscapes dataset (Cordts et al., 2016), our tomato plant disease dataset and several other datasets provided in (Isola et al., 2017; Zhu et al., 2017). The tomato plant disease dataset is also used to investigate the effectiveness of synthetic data augmentation in the recognition system of our proposed method.

### 3.1.1. Tomato Plant Disease Dataset

We collected 2,789 tomato plant images with 9 identifiable disease classes (Fig. 3(ii)) for our dataset under different circumstances depending on the illumination, season, temperature, humidity, and places where they were taken using simple camera devices. For that purpose, we have supported our dataset with images having distinct features and conditions like (i) Samples at different infection status. (ii) Varied sizes of the plant. (iii) Images containing different infected regions of the plant (e.g., stem, leaves, fruits, etc.). (iv) Images with diverse resolutions. (v) Objects surrounding the plant, etc. These conditions help to approximate the infection process and deduce how a plant is affected by the pest or disease (origin or possible developing cause).

After collecting images, we manually annotate by capturing the RoIs in the samples as the main part of images and categorizing each image containing the disease or pest into its respective class, as shown in Fig. 3(i). The opinion from the experts in the area was a must, depending on the infection status in different diseases that look similar, where the knowledge for identifying the disease was required. This serves as our ground truth.

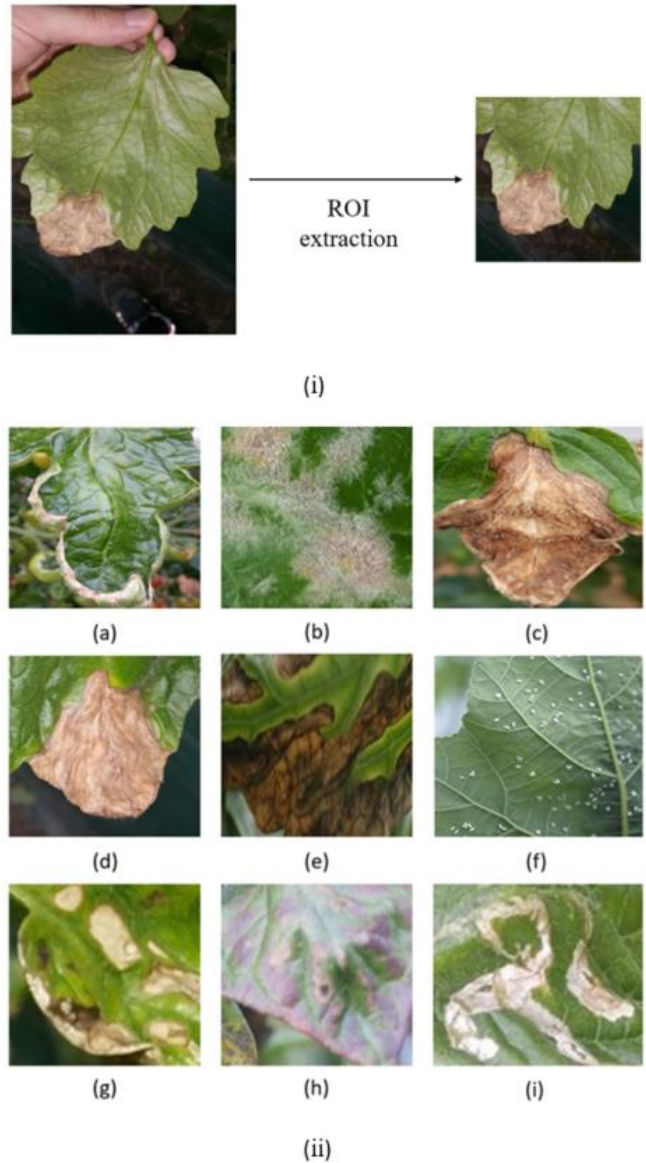

**Fig. 3.** ROI extraction (i) RoI sample extraction process. (ii) Examples of extracted ROIs with diseases and pests affecting tomato plants from our tomato plant disease dataset. (a) Nutritional excess or deficiency, (b) Powdery mildew, (c) Gray mold, (d) Plague, (e) Canker, (f) Whitefly, (g) Leaf mold, (h) Low temperature, (i) Miner.

### 3.2. Metrics

We adopt FCN (Isola et al., 2017), FID (Heusel et al., 2017) and NIMA (Talebi and Milanfar, 2018) as the evaluation criterion for our generative model.

3.2.1. FCN score

We use FCN score (Isola et al., 2017) for evaluating our generative model as an automatic quantitative measure that does not require any human experiments. We adopt this metric to evaluate on the Cityscapes labels→photo task. The FCN metric evaluates the comprehensibility of the generated photos in accordance to a semantic segmentation algorithm (the fully-convolutional network, FCN [(Long et al., 2015)]). The FCN predicts a label map for the generated photos. The label maps are then compared to the input ground truth labels using standard semantic segmentation metrics like mean Intersection-Over-Union (Class IoU), per-class accuracy (pca), and per-pixel accuracy (ppa).

3.2.2. FID score

We also evaluate our results with the metric Fréchet Inception Distance (FID) score [(Heusel et al., 2017)] that correlates well with human judgment. FID score is known to capture the similarity of generated images to real ones better than the Inception Score (Salimans et al., 2016). Smaller FID value denotes better quality. Even though this metric allows us to avoid depending on human evaluations as they associate well with our subjective judgment of image quality (Salimans et al., 2016), it is recommended to use a huge sample size to calculate this metric otherwise the true realization of the generator is underestimated. For our tomato plant disease dataset, we use 10,564 random crops of healthy tomato plant leaves and translate them to leaves infected with powdery mildew to evaluate using FID score. Considering that there are not many samples in the cityscapes dataset to evaluate and easily tell the difference, our results are mainly based on manual inspection of the visual fidelity of the generated images.

3.2.3. NIMA score

Apart from the metrics mentioned above, we also evaluate our model using the state-of-the art evaluation model namely neural image assessment (NIMA) (Talebi and Milanfar, 2018). The NIMA estimates aesthetic qualities in aspects of photography skills and perceptual relevance. Larger values of NIMA denote better aesthetic quality of an image. We calculate the average of the NIMA score for the images in the test set of the datasets used for evaluation of the generative model.

We adopt the class precision and total accuracy metrics as the evaluation criterion for our recognition task. Accuracy has often been adopted as a major metric to evaluate recognition algorithms while the example-based evaluation metric like class precision is better at capturing the consistency of predictions on a given diseased class image.

## 3.3. Results

We compare the results of AR-GAN with various approaches including supervised pix2pix (Isola et al., 2017). We present sample results from AR-GAN across given datasets to demonstrate the performance in effective translation of features between two domains. Further, we analyze our recognition CNN to evaluate the performance of classical and synthetic augmentation using various instances of our tomato plant disease data.

### 3.3.1. Comparing ARL-GAN against other GAN models

We first compare our AR-GAN against several recently proposed models for unpaired image-to-image translation on paired cityscapes dataset with 2975 training images. To perform the image-to-image translation task, all the

Table 1. FCN-scores for different methods, evaluated on Cityscapes labels→photos.

| Loss | Per-pixel accuracy | Per-class accuracy | Class IoU |
|---|---|---|---|
| CoGAN [9] | 0.40 | 0.10 | 0.06 |
| BiGAN [26] | 0.19 | 0.06 | 0.02 |
| SimGAN [3] | 0.20 | 0.10 | 0.04 |
| CycleGAN [32] | 0.52 | 0.17 | 0.11 |
| ARL-GAN | **0.68** | **0.20** | **0.15** |
| pix2pix* [18] | 0.71 | 0.25 | 0.18 |
| Cityscapes test set* | 0.80 | 0.26 | 0.21 |

models are required to capture the semantic information from the input image and generate the corresponding transformed image. For a fair comparison as reported in Table 1, all the methods use the same architecture and details as reported in (Zhu et al., 2017). Here, our AR-GAN uses 9 residual blocks and Activation Reconstruction Loss in the forward direction:

$\mathcal{L}_{ARL_f}$. We assess their performance on Cityscapes labels→photo task and observe that our method outperforms them in terms of FCN metrics. We observe that the CycleGAN model based on cycle reconstruction and adversarial loss can be improved with our proposed addition of ARL between the real and fake image pairs. The translations produced using our approach are often of similar quality to the fully supervised pix2pix.

### 3.3.2. Comparing AR-GAN against CycleGAN variants

In Tables 2 and 3, we evaluate different generator architectural variants and compare their performance in terms of FID and NIMA using powdery mildew class from the tomato plant disease dataset and cityscapes dataset, respectively. We also evaluate our method with the ARL in both directions: $\mathcal{L}_{ARL_f}$ between $a$ and $G_{AB}(a)$ and $\mathcal{L}_{ARL_b}$ between $b$ and $G_{BA}(b)$. In the tables, we use res*, where * denotes the number of residual blocks in the generator architecture. Last model in Table 2 uses AR-GAN with 9 residual blocks in the generator and ARL in both directions: $\mathcal{L}_{ARL_f} + \mathcal{L}_{ARL_b}$. Our method outperforms all these variants with a significant difference between the resulting scores. Figs. 4 and 5 show samples generated from generators of different models in Tables 2 and 3 on powdery mildew class from the tomato disease dataset. We are unable to achieve compelling results with CycleGAN. AR-GAN, on the other hand, can produce translations with higher fidelity and aesthetic value. It generates compositions that are not only more faithful to the input, but also have fewer artifacts. CycleGAN often confuses between the foreground and the background, while AR-GAN explicitly deals with it by getting sharper edges of the target, without treating the whole image as one object. As can also be seen from Table 1, CycleGAN finds it hard to capture perceptual information from the input image in the cityscapes dataset, resulting in a poor FCN score, when compared to AR-GAN.

**Table 2.** FID and NIMA scores on healthy tomato leaves→leaves infected with powdery mildew at 256 × 256 resolution.

| Model | FID (lower is better) | NIMA (higher is better) |
|---|---|---|
| CycleGAN (res6) | 75.81 | 4.842 ± 1.841 |
| CycleGAN (res9) | 92.26 | 4.584 ± 1.860 |
| CycleGAN (res12) | 88.86 | 4.804 ± 1.831 |
| CycleGAN (U-net[31]) | 86.16 | 4.612 ± 1.859 |
| AR-GAN( res9, $\mathcal{L}_{ARL_f}$) | 55.18 | 4.894 ± 1.806 |
| AR-GAN(U-net[31], $\mathcal{L}_{ARL_f}$) | **47.15** | 4.909 ± 1.803 |
| AR-GAN( res9, $\mathcal{L}_{ARL_f} + \mathcal{L}_{ARL_b}$) | 54.65 | **4.911 ± 1.809** |

**Table 3.** FID and NIMA on cityscapes labels→photos at 256 × 256 resolution.

| Model | FID (lower is better) | NIMA (higher is better) |
|---|---|---|
| CycleGAN (res6) | 66.95 | 4.639 ± 1.697 |
| CycleGAN (res9) | 74.25 | 4.588 ± 1.672 |
| CycleGAN (res12) | 66.68 | 4.623 ± 1.681 |
| AR-GAN( res9, $\mathcal{L}_{ARL_f}$) | **56.23** | **4.643 ± 1.652** |

### 3.3.3. Analysis of loss function

In image translation methods, the fidelity and resemblance between images generated by GANs and real images is a trade-off. In our model, it is determined by the weight λ of the ARL relative to the cycle-adversarial term. In our experiments, we start by setting λ = 0 and gradually increase its value in regular increments until a threshold is reached where the translated image looks closer to the input but fails to pertain the visual traits from the target domain. This effect is demonstrated in Fig. 6, on horse→zebra translation task on unpaired Horse2Zebra dataset (Zhu et al., 2017). Tuning λ for each specific task and dataset gives results that are both faithful to the original image and similar to the target domain, and thus penalizes the perceptual quality between the images from the two domains.

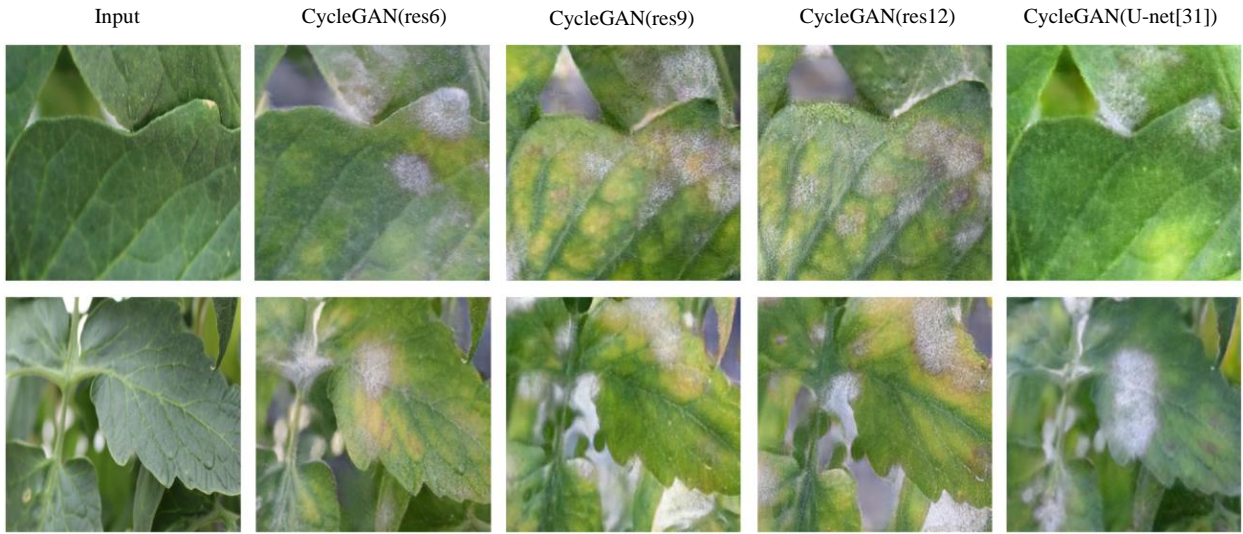

**Fig. 4.** Different variants of CycleGAN for mapping healthy leaves→leaves infected with powdery mildew from our unpaired tomato plant disease data. From left to right: input, CycleGAN with resnet6 generator, CycleGAN with resnet9 generator, CycleGAN with resnet12 generator and CycleGAN with Unet generator.

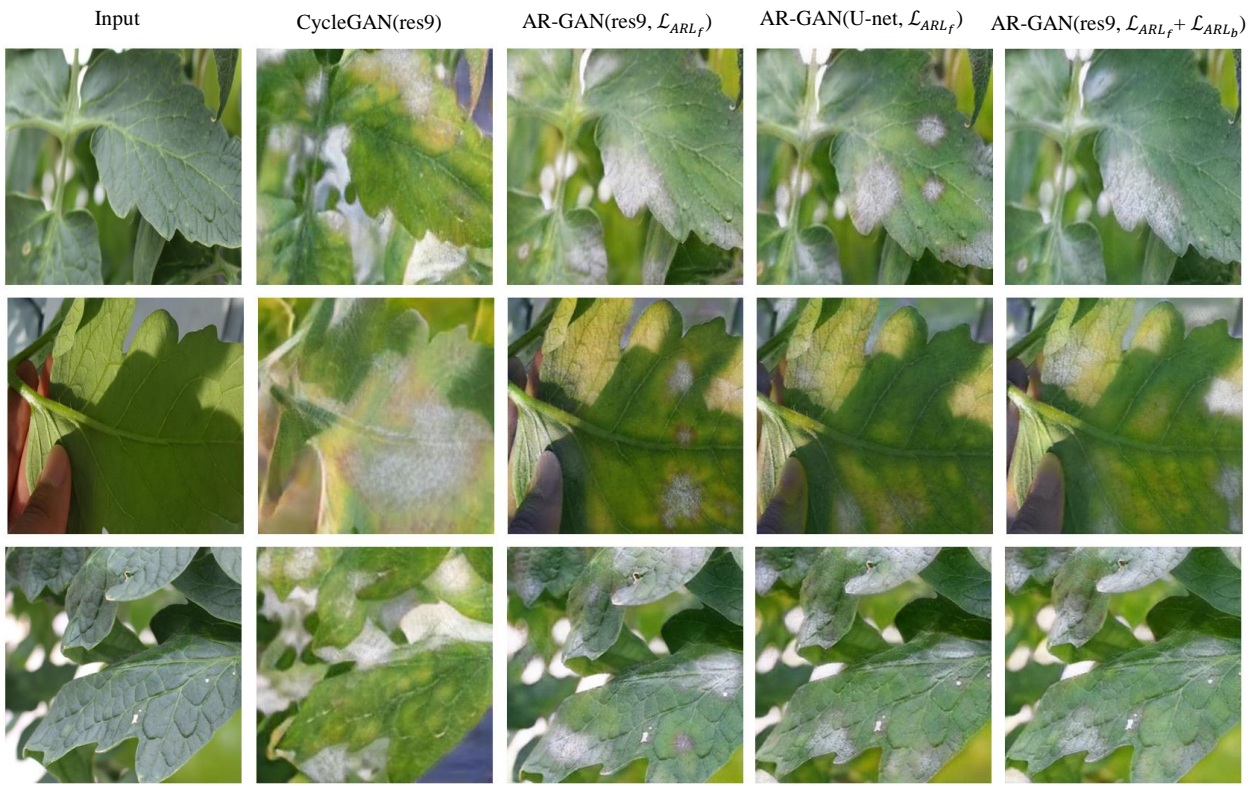

**Fig. 5.** Different variants of our method for mapping healthy leaves→leaves infected with powdery mildew from our unpaired tomato plant disease data. From left to right: input, CycleGAN with resnet9 generator, AR-GAN with resnet9 generator, AR-GAN with Unet generator, AR-GAN with Unet generator and ARL between real and fake images in both directions: $\mathcal{L}_{ARL_f} + \mathcal{L}_{ARL_b}$.

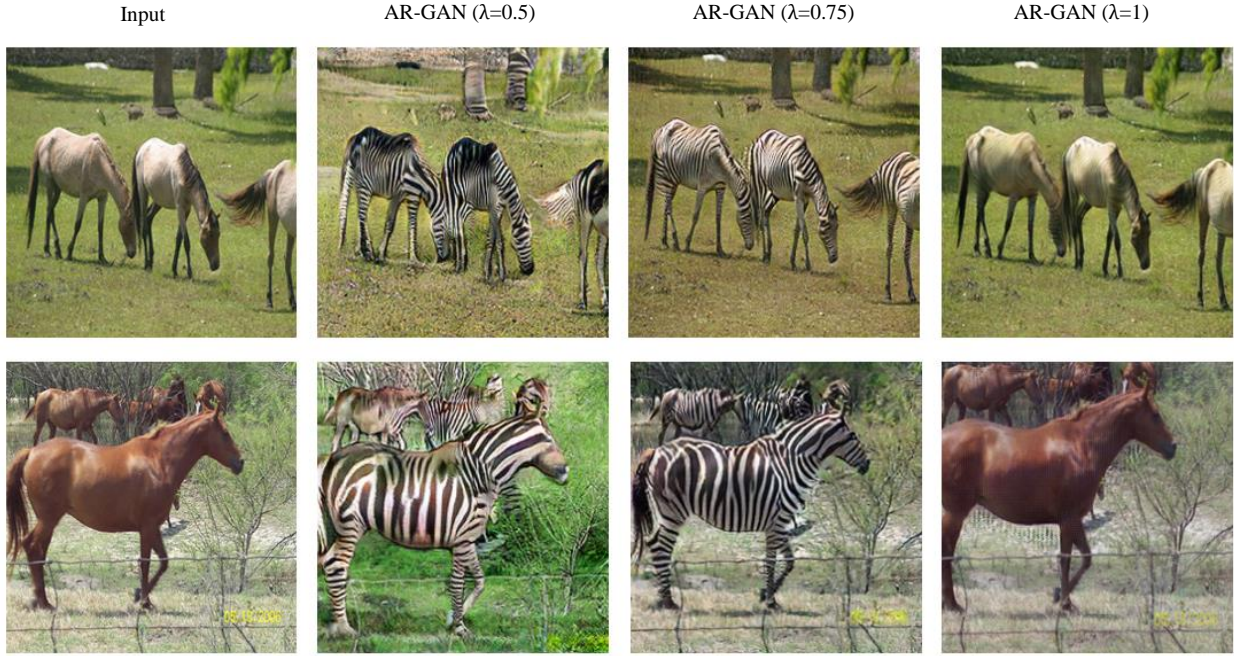

**Fig. 6.** Mapping horses→zebras using 1171 images from zebra class and 939 images from wild horse class of ImageNet. From left to right: input, AR-GAN ($\lambda$=0.5), AR-GAN ($\lambda$=0.75) and AR-GAN ($\lambda$=1). While $\lambda$=0.75 gives the best results for the Horse2Zebra dataset, in our experiments with Tomato Plant Disease Dataset we find it suitable to use $\lambda = 1$, which depicts $\lambda$ as a hyperparameter depending on the nature of data in the two domains while training the model.

### 3.3.4. Data augmentation

After extracting the RoIs, the number of plant disease images in our dataset is 5,318 with class distribution portrayed as $X$ in Fig. 7. As evident from Fig. 7, the available limited training data $X$ suffers from heavy class imbalance. Among 9 classes, two classes of 'Leaf Mold' and 'Whitefly' occupy over 50% of the dataset while 'Low Temperature' and 'Powdery Mildew' occupy less than 5%. To address this problem of data insufficiency and heavy class imbalance, we augment dataset $X$ using two kinds of data augmentation methods: classical data augmentation and synthetic data augmentation using AR-GAN. Through these data augmentation processes, the numbers of data instances can be increased by more than double, for classes with minimum samples. It can also be seen from Fig. 7 that using data augmentation, we have a relatively better balance between classes. In Fig. 7, $X$ represents the original dataset obtained after RoI extraction, $X + X_C$ represents the dataset augmented by classical augmentation of $X$, and $X + X_S$ represents dataset augmented by synthetic augmentation of $X$. Note that the number of samples per class in $X + X_C$ and $X + X_S$ is kept proportional for fair evaluation. We do not use any augmented data for classes 'Leaf Mold' and 'Whitefly' that means $X$, $X + X_C$ and $X + X_S$ for these two classes are the same.

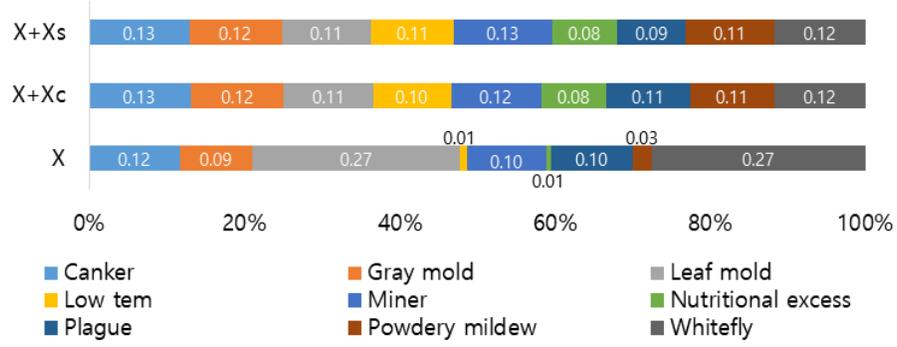

**Fig. 7.** Class distributions of Tomato Plant Disease Dataset instances under different data augmentation methods, (i) $X$: Original dataset after RoI extraction, (ii) $X + X_C$: Dataset augmented classically from $X$, and (iii) $X + X_S$: Dataset augmented synthetically from $X$.

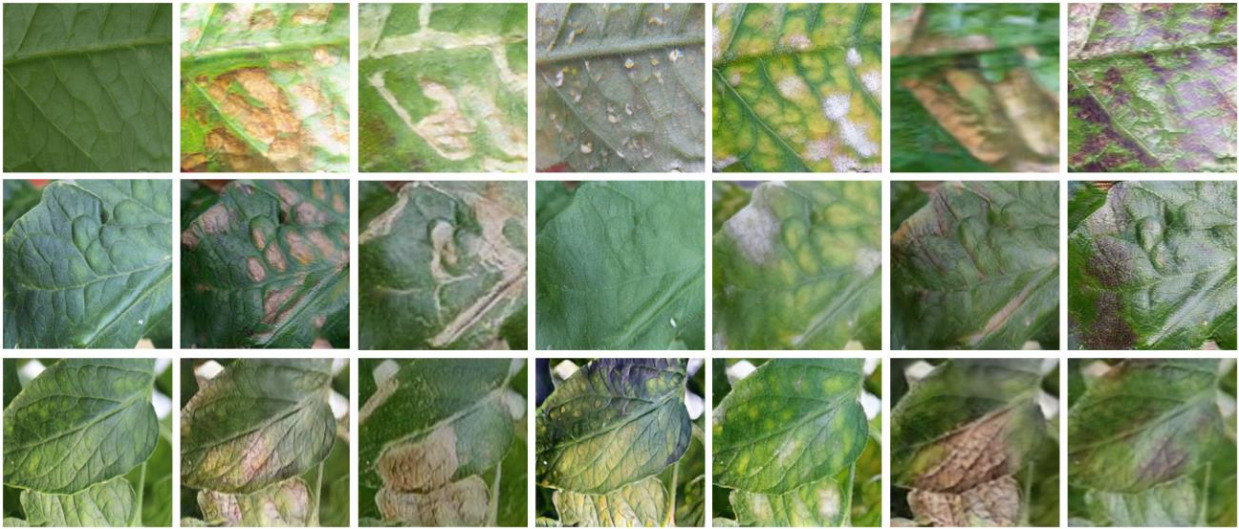

**Fig. 8.** Synthetic images from AR-GAN model. The real images shown at the leftmost column are inputs that the synthetic images are based on. The 2nd, 3rd, 4th, 5th, 6th and 7th column represent the translated images to the domains: canker, miner, whitefly, powdery mildew, plague and low temperature respectively.

Fig. 8 shows samples of images generated by our AR-GAN for each class. For classical augmentation, the extracted RoIs are first induced with random distortions like tilting, skewing and shearing to produce $X_{dist}$ samples with a probability of 1 each. Each image is also rotated $X_{rot}$ times at random angles with θ = [0°, ..., 180°] with probability of 1 and flipped up-down and left-right $X_{flip}$ times with probability of 0.5. Elastic augmentation, $X_{ed}$ is then applied with a grid size of 16 and probability of 1 to all the resultant images. An example of an extracted RoI and its corresponding augmentation are shown in Figure 3.3. This augmentation process results in a total number of augmentations $X_c$, such that:

$$X_c = (X_{rot} + X_{dist} + X_{flip} + X) \times X_{ed} \quad (7)$$

All the resulting samples were resized to 256 × 256 using bicubic interpolation.

### 3.3.5. Analysis of classic and synthetic augmentation techniques

To analyze the results of our baseline recognition CNN - ResNet50 and show the performance of classical and synthetic data augmentation, we use three instances of our tomato plant disease dataset ($X, X + X_C, X + X_S$) as training data. To evaluate the recognition system, we use 200 test samples from each class, which remain unseen to this recognition system while training. From Fig. 9, we observe that using classical augmentation to increase the size of the training set, shows an increase from 80.9% to 81.7% (+0.8) in terms of accuracy. Compared to classical augmentation, our method of synthetically generating tomato plant disease images to aid the training set shows an increase from 80.9% to 86.1% (+5.2%).

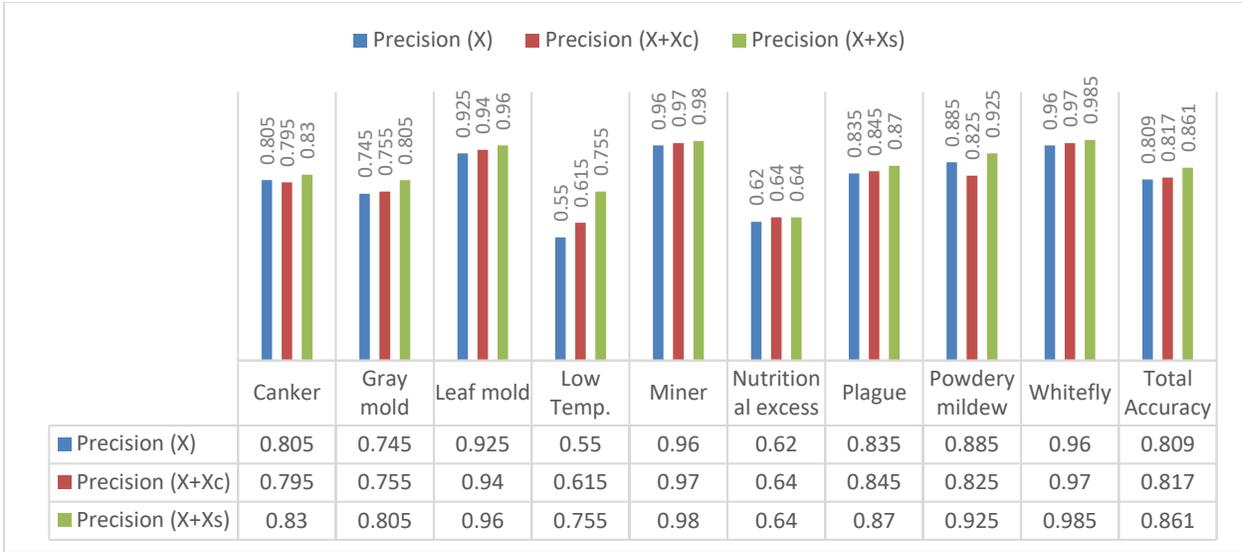

|  | Canker | Gray mold | Leaf mold | Low Temp. | Miner | Nutritional excess | Plague | Powdery mildew | Whitefly | Total Accuracy |
|---|---|---|---|---|---|---|---|---|---|---|
| Precision (X) | 0.805 | 0.745 | 0.925 | 0.55 | 0.96 | 0.62 | 0.835 | 0.885 | 0.96 | 0.809 |
| Precision (X+Xc) | 0.795 | 0.755 | 0.94 | 0.615 | 0.97 | 0.64 | 0.845 | 0.825 | 0.97 | 0.817 |
| Precision (X+Xs) | 0.83 | 0.805 | 0.96 | 0.755 | 0.98 | 0.64 | 0.87 | 0.925 | 0.985 | 0.861 |

**Fig. 9.** Class Precision and total accuracy of the baseline recognition CNN at dataset instances: $X, X + X_C, X + X_S$ of tomato plant disease dataset. We observe a significant improvement in class and final average precision from $X \rightarrow X + X_C \rightarrow X + X_S$. The final column represents the total accuracy for each data instance on the validation data.

For further analysis, we investigate the changes in predicted results of 200 positives from each class used to test our recognition system using the three different training dataset instances: $X, X + X_C$, and $X + X_S$. Fig. 10 shows relative changes of predicted true positives of two augmented datasets $X + X_C$, and $X + X_S$ based on $X$. For all classes, using synthetically augmented dataset $X + X_S$ as a training dataset, increases the portion of true positives compared to the original dataset $X$. However, in the case of training with dataset instance $X + X_C$, it shows worse results for classes 'Canker' and 'Powdery mildew'. The average rate of increase in the portion of true positives remains much higher when using synthetically augmented dataset instance as compared to the dataset instance using classic augmentation techniques.

We also compare the relative change rate of false positives for each tomato plant disease when using data instances $X + X_C$ and $X + X_S$ against the original dataset in Fig. 11 and illustrate the detailed changes in the false positive rates leading to the improvement in the final accuracy. Three horizontal bars in each row of the corresponding class represent the ratios of the amount of false positives between various instances of our dataset. The first two bars represent the ratio of amount of false positives using dataset instances $X + X_c$ and $X + X_s$ to the amount of false positives from the original dataset $X$, respectively. The third bar represents the ratio of false positives using dataset instances $X + X_s$ and $X + X_c$. A negative rate displayed on the left side of the graph represents the decrease in false positives while the positive rate displayed on the right side represents the increase in false positives of the corresponding class. Comparing dataset instances $X + X_s$ and $X + X_c$, as represented in the third horizontal bar for each row, we can identify longer stacked negative rate bars than positive ones for all classes resulting in an improved overall accuracy. As can be seen from Fig. 10, there is a significant increase in the relative change of rate in true positives, especially in classes gray mold, low temperature and powdery mildew. This effect can also be seen in Fig. 11 showing that these classes have relatively more decrease in the relative rate of change in false positives. We can also observe that the classes with a relative increment in class distribution (as shown in Fig. 7) due to data augmentation, e.g. 'Gray mold', 'Low temperature', 'Nutritional excess', and 'Powdery mildew' show relatively exorbitant changes in false positives. These classes show more negative changes (reduction of false positives) than the other classes when using synthetic augmentation. This comparative illustration shows that the diversity and variation that can be gained from classical modifications of the images for augmentation is relatively smaller when compared to synthetic augmentation. These results also indicate that using generated tomato plant disease images together with real images yield improvements in the performance for the recognition task over the limited tomato plant disease data.

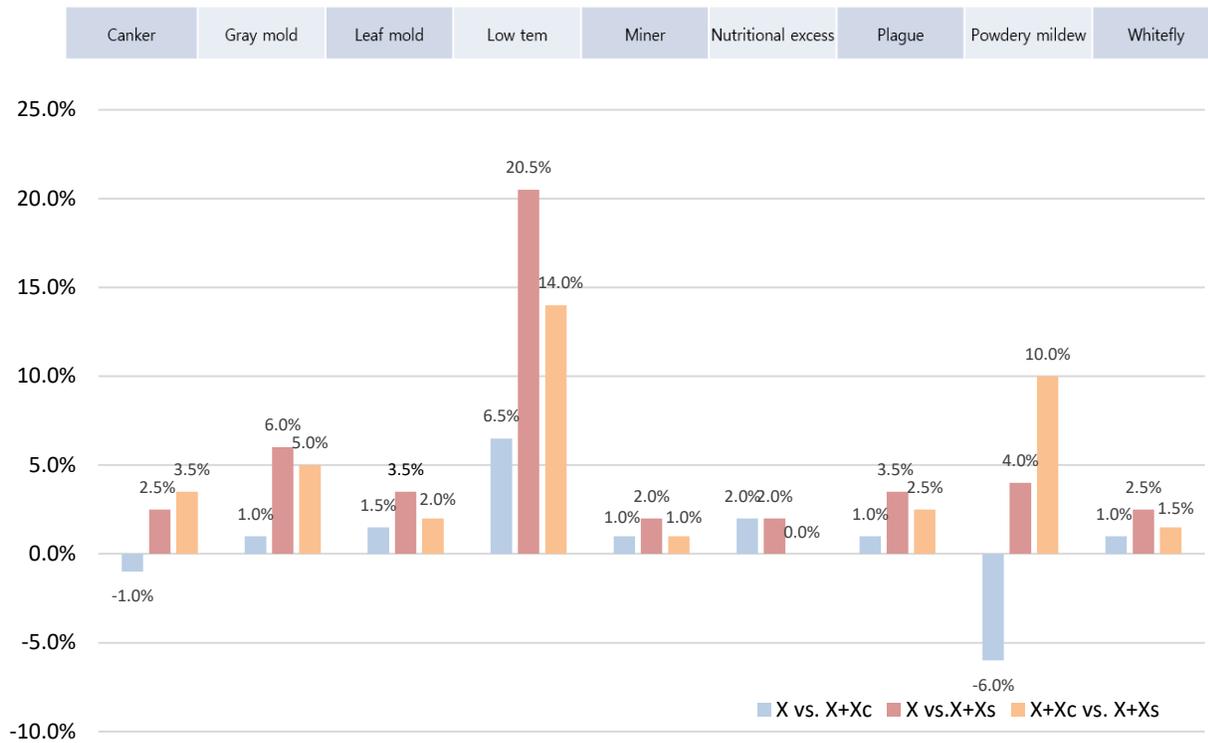

**Fig. 10.** Relative change rate of true positives for two augmented training datasets $X + X_C$, and $X + X_S$ against the original dataset $X$.

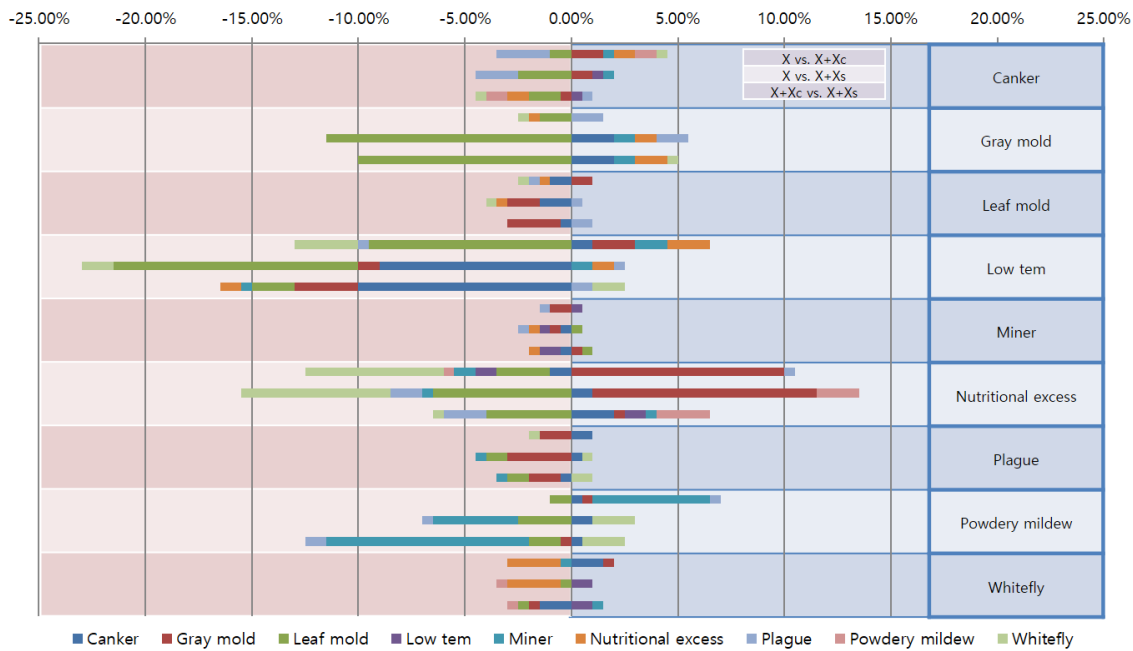

**Fig. 11.** Relative change rate of false positives for each tomato plant disease when using two training dataset instances $X + X_C$, and $X + X_S$ against the original dataset $X$.

## 4. Discussion

This work focused on using GANs in an image translation setting to synthetically augment plant disease dataset and further improve the performance on the plant disease recognition task using deep CNN. Our relatively small dataset reflects the size of datasets available to most researchers in the field of plant disease detection and recognition. We first extend the state-of-the-art for GAN-based image-to-image synthesis to improving the perceptual quality of the generated images. In addition to a cycle consistent and adversarial term, our GAN based image-to-image translation framework (AR-GAN) optimizes on Activation Reconstruction loss function that measures feature activation against the real image. We present visually more convincing synthetic images in comparison to one of the most prominent existing models and evaluate the performance of our AR-GAN framework in terms of different datasets and metrics. Further, we tested our hypothesis that adding synthetic samples would improve classification results. We analyze a baseline convolutional neural network classifier for improved recognition using the synthetic samples for augmentation of our training set and compare it with the classical data augmentation scheme. We observe a significant improvement in classification accuracy (+5.2%) using the synthetic samples generated by AR-GAN framework as compared to (+0.8%) increase using classic augmentation strategy.

Our results show that the images generated using AR-GAN have meaningful features and visualizations that can be incorporated into other computer aided algorithms. In the future, we plan to extend our work to additional computer vision domains that can yield better performance using synthetic plant disease generation. We believe that problems like detection and recognition of plant diseases can benefit from using synthetic augmentation, and that the presented approach can lead to a stronger and more robust plant disease detection systems.

## Acknowledgements

This research was supported by Basic Science Research Program through the National Research Foundation of Korea (NRF) funded by the Ministry of Education (No. 2019R1A6A1A09031717). This work was carried out with the support of "Cooperative Research Program for Agriculture Science and Technology Development (Project No. PJ01389105)" Rural Development Administration, Republic of Korea.